# A Convolutional Neural Network with Parallel Multi-Scale Spatial Pooling to Detect Temporal Changes in SAR Images

Jia-Wei Chen 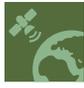, Rongfang Wang *, Fan Ding, Bo Liu, Licheng Jiao and Jie Zhang

Key Laboratory of Intelligent Perception and Image Understanding of Ministry of Education, School of Artificial Intelligence, Xidian University, Xi'an 710071, China; jawaechan@gmail.com (J.-W.C.); dfmamba@163.com (F.D.); liub@xidian.edu.cn (B.L.); lchjiao@mail.xidian.edu.cn (L.J.); j_zhang@stu.xidian.edu.cn (J.Z.)

* Correspondence: rfwang@xidian.edu.cn

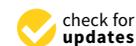

**Abstract:** In synthetic aperture radar (SAR) image change detection, it is quite challenging to exploit the changing information from the noisy difference image subject to the speckle. In this paper, we propose a multi-scale spatial pooling (MSSP) network to exploit the changed information from the noisy difference image. Being different from the traditional convolutional network with only mono-scale pooling kernels, in the proposed method, multi-scale pooling kernels are equipped in a convolutional network to exploit the spatial context information on changed regions from the difference image. Furthermore, to verify the generalization of the proposed method, we apply our proposed method to the cross-dataset bitemporal SAR image change detection, where the MSSP network (MSSP-Net) is trained on a dataset and then applied to an unknown testing dataset. We compare the proposed method with other state-of-arts and the comparisons are performed on four challenging datasets of bitemporal SAR images. Experimental results demonstrate that our proposed method obtain comparable results with S-PCA-Net on YR-A and YR-B dataset and outperforms other state-of-art methods, especially on the Sendai-A and Sendai-B datasets with more complex scenes. More important, MSSP-Net is more efficient than S-PCA-Net and convolutional neural networks (CNN) with less executing time in both training and testing phases.

## 1. Introduction

Synthetic aperture radar (SAR) is a microwave sensor for earth observation working without the limitations of illumination condition. This advantage allows people to perform multiple earth observations at all time with all weather and the acquired multitemporal SAR images give us opportunities to compare the difference of the multi-temporal SAR images on the same scene, which is known as multi-temporal SAR image change detection [1]. In recent years, numerous methods have been developed for SAR image change detection.

Currently, most SAR image change detection methods are developed based on the framework proposed in [2,3] by L. Bruzzone and D. F. Prieto, in which the changed regions are detected from a difference image (DI). However, this pixel-wise operator is subject to SAR image speckle and it is quite challenging to exploit the changing information. Zhang et al. [4] proposed a graph-cut method to extract the change regions on the log-ratio difference image through the statistical distributions on the changed and unchanged regions. Li et al. [5] proposed a joint sparse learning model to obtain robust features from difference images.

Recently, several fully convolutional neural networks (CNN) [6] have been successfully employed to image semantic segmentation where the pooling layers can exploit robust features on spatial structures of an image. Especially, three-dimension (3-D) CNNs have been proposed for 3D image and employed to remote sensing analysis [7,8] and combined with transfer learning [9,10]. Inspired by this spirit, they have been extensively employed to exploit the changed regions from the noisy difference image. Gong et al. [11] proposed a deep neural network for the first time to SAR image change detection. Gao et al. [12] proposed a simple convolutional network based on the principle component analysis, known as PCA-Net, to SAR image change detection. Li et al. [13] proposed a bitemporal SAR image change detection based on convolutional neural network. However, most above learning-based methods are unsupervised ones or the models are trained by pseudo labels estimated by an intermedia model. Then the performance of a trained model is limited by the errors of pseudo labels accumulated in iterations. To handle this problem, Wang et al. [14] proposed a supervised PCA-Net approach, where training samples are selected with the guidance of morphological structures of reference. However, the PCA-Net is time-consuming to train a promising model. For another, in most traditional convolutional networks, all the pooling kernels have the same size and pooling operators are usually subsequently employed to exploit a larger range of spatial context. Recently, Zhao et al. [15] proposed a pyramid scene parsing network, which exploits global spatial context information by aggregating various sizes of context through pyramid pooling layers. Kim et al. [16] developed an U-Net with pyramid pooling layers for object segmentation. Cui et al. [17] proposed a multi-scale SAR image segmentation based on attention-based CNN.

In this paper, our goal is to build an efficient and easy-training network for SAR image change detection. To achieve this, we propose a light multi-scale spatial pooling (MSSP) convolutional network (MSSP-Net) to segment the changed regions from the noisy difference image. In this network, the MSSP layer is introduced to obtain the robust features of changed regions with the spatial context information at various scales. It can facilitate to exploit the structures of changed regions with a more shallow network, which can be easily trained as a low capacity model. Furthermore, We also focus on the generalizing ability of the proposed method. Besides the experiments on change detection of individual dataset, we verify the model generalization by comparing the methods on the cross-dataset SAR image change detection. The experimental results show that the proposed method outperforms the other methods in cross-dataset SAR image change detection, especially on the datasets with complex scenes. Compared with the existing CNN-based change detection method, the proposed MSSP-Net can achieve multi-scale receptive field with a shallow and light-weighted convolutional neural network.

## 2. Proposed Method

In most traditional convolutional neural network (CNN), the pooling operators with mono-scale are usually employed to exploit the context information. However, the exploitation at various scales is usually achieved by cascading these operators and convolutional layers, and it will increase the depth of network and the number of model parameters. In this section, we propose an efficient covolutional neural network with a multiscale spatial pooling layer to exploit the structures of changed regions from bitemporal SAR images.

*2.1. Network Architecture*

Given a set of bitemporal SAR images, the DI is firstly generated by the neighborhood-based log-ratio(LR) ratio [18]. Then both the bitemporal SAR images and the generated DI are taken as the input of the MSSP network. The goal is to elaborate the changed map from the noisy input images.

The whole framework of MSSP network (MSSP-Net) is illustrated in Figure 1. We take three patches with size $32 \times 32$ as inputs and list the sizes of all the intermedia activation tensors in Table 1.

To perform efficient inference, we extract patches from the group of input images and feed the patches into the MSSP network. As shown in Figure 1, for each group of the input patches, it is sequentially followed by a batch-norm (BN) layer and three convolution layers (Conv-3), illustrated as the blue bars in the figure. The digit in 'Conv-3' indicates the size of the convolution kernels is $3\times3$.

Then a max-pooling layer (MP-2) follows the convolution layers to exploit spatial context with a 2×2 kernel as illustrated by the red bar.

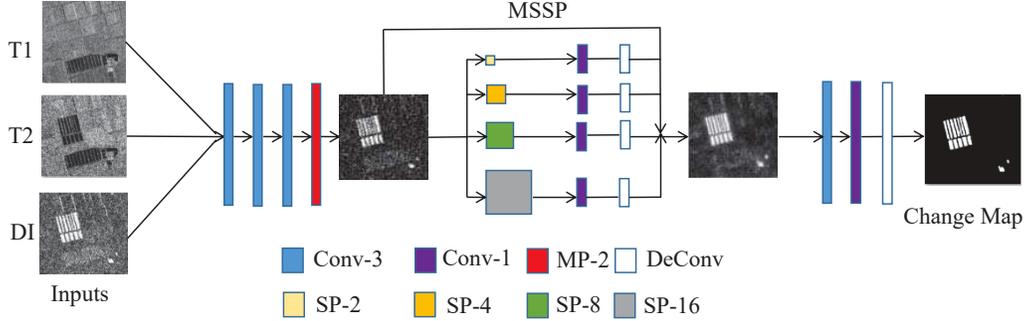

**Figure 1.** The illustration of the multi-scale spatial pooling (MSSP)-net.

**Table 1.** The architecture of MSSP-net.

| Layer | Tensor | Layer | Tensor | Layer | Tensor |
|---|---|---|---|---|---|
| Input | $32 \times 32 \times 3$ | | | | |
| Conv-3.1 | $32 \times 32 \times 32$ | Conv-3.2 | $32 \times 32 \times 64$ | Conv-3.3 | $32 \times 32 \times 128$ |
| MP-2 | $16 \times 16 \times 128$ | | | | |
| SP-2 | $8 \times 8 \times 128$ | Conv-1.1 | $8 \times 8 \times 32$ | DeConv.1 | $16 \times 16 \times 32$ |
| SP-4 | $4 \times 4 \times 128$ | Conv-1.2 | $4 \times 4 \times 32$ | DeConv.2 | $16 \times 16 \times 32$ |
| SP-8 | $2 \times 2 \times 128$ | Conv-1.3 | $2 \times 2 \times 32$ | DeConv.3 | $16 \times 16 \times 32$ |
| SP-16 | $1 \times 1 \times 128$ | Conv-1.4 | $1 \times 1 \times 32$ | DeConv.4 | $16 \times 16 \times 32$ |
| Contact | $16 \times 16 \times 256$ | Conv-3.4 | $16 \times 16 \times 64$ | Conv-1.5 | $16 \times 16 \times 2$ |
| DeConv.5 | $32 \times 32 \times 2$ | | | | |

Following the MP-2 layer, a MSSP layer is developed to parallelly exploit the spatial context with various scales of receptive fields. To achieve this, the MSSP layer is designed as a group of parallel convolutional layers with the kernel sizes of 2×2, 4×4, 8×8 and 16×16 illustrated by various sizes and colors of blocks and denoted by SP-2, SP-4, SP-8 and SP-16, respectively.

Being different from the conventional CNN, our proposed MSSP layer parallelly exploits the spatial context with pooling layers at various scales, which reduces the depth of neural network and makes it easier to train.

After each spatial pooling operator, a convolutional and a deconvolutional operator are employed to perform the upsampling that recovers the patch to the size of $16 \times 16$, where the spatial context obtained by the MSSP will be propagated to pixel levels [19].

The output of the MSSP layer can be expressed as follows

$$\mathbf{I}_{out} = \bigoplus_{i=1,2,\ldots} \mathbf{I}_{in} \star \mathbf{a}_i \star \mathbf{f}_i, \tag{1}$$

where $\mathbf{I}_{in}$ and $\mathbf{I}_{out}$ are the input and the output of MSSP layer. $\mathbf{a}_i$ is the i-th spatial pooling operator with the scale of $s_i$. In this paper, we set $s_i = 2, 4, 8, 16$. $\mathbf{f}_i$ is the i-th deconvolutional kernel. $\bigoplus$ denote the direct sum of feature subspace generated by various $\mathbf{a}_i$. $\star$ denotes the convolution operator. In here, the direct sum $\bigoplus$ is implemented as a contact operator. Namely, we contact the output of MSSP layer and its input. Then we get a tensor with size $16 \times 16 \times 256$. Next, a 3×3 and a $1 \times 1$ convolution are employed to get two probability maps for changed and unchanged categories. Finally, we employ deconvolution again to upsample the probability map to the same size of input patch.

As shown above, the MSSP-Net is a lite and shallow network and most importantly, we exploit the spatial context at the various scales by a MSSP layer instead of cascading mono-scale pooling layers, which reduces the depth of the network and improves the computational efficiency.

*2.2. Network Training*

To train the network efficiently, we elaborately collected the samples from training datasets, especially near the boundaries (named as boundary samples) according to the method in [14]. It has been demonstrated that the network can be efficiently trained for SAR image change detection with less training samples. In this paper, we randomly draw 20% samples for training including changed and unchanged category, among which there are 50% samples along boundaries.

Moreover, the patchsize of each sample was set as $32 \times 32$ and 8 samples were fed at each training step. Additionally, the Adam algorithm [20] was employed to optimize the weights of network in the training stage, where the initial learning rate was set as 0.005. Training was performed on the PyTorch platform built on the Ubuntu 16.04 installed in a PC with a 16 GB DDR memory and an NVIDIA TITAN Xp Graphics Processing Unit of 11 GB memory.

**3. Experimental Results and Analysis**

*3.1. Datasets Description and Experiment Configures*

In this paper, the proposed method is verified on four sets of bitemporal SAR images. Two scenes (YR-A and YR-B) are from bitemporal Yellow River SAR images [11] acquired by the Radarsat-2 satellite in 2008 and 2009, respectively. Their image sizes are $306 \times 291$ and $400 \times 350$, respectively. We manually labeled their ground truth. Other two are parts of TerraSAR-X images acquired prior to (on Oct. 20, 2010) and after (on May 6, 2011) the Sendai earthquake in Japan [21]. Their sizes (Sendai-A and Sendai-B) are $590 \times 687$ and $689 \times 734$, respectively. Their reference maps are kindly provided by [21]. These four datasets are shown in Figure 2, where the images in the first two rows are the bitemporal SAR images and the images in the last row are the corresponding DIs. These four datasets are quite challenging. As shown in the last row of Figure 2, the changed regions are subject to the heavy noise. More specially, in both Sendai-A and Sendai-B datasets, the scenes of non-changed area are quite complex.

In the following experiments, the performance will be quantitatively evaluated by probabilistic missed alarm (pMA), probabilistic false alarm (pFA) and kappa coefficient, where pFA (pMA) are calculated by the ratios between FA (MA) and the number of unchanged pixels (NC).

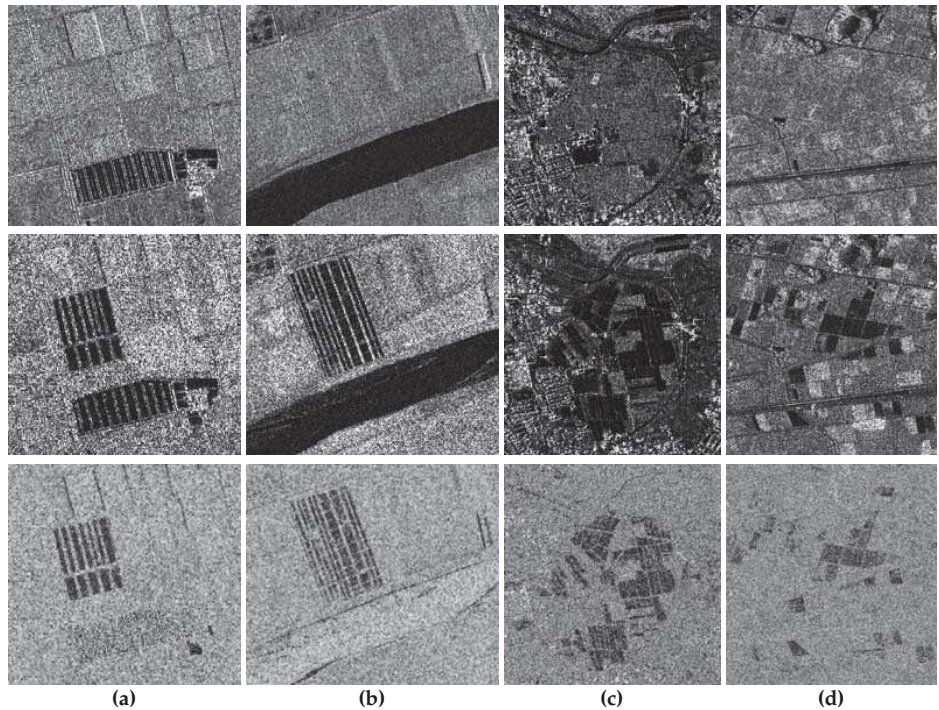

**Figure 2.** Four sets of bitemporal synthetic aperture radar (SAR) images: (**a**) YR-A, (**b**) YR-B, (**c**) Sendai-A, and (**d**) Sendai-B. The images in first two rows are bitemporal SAR images and the last row are the difference images (DIs).

*3.2. Experiment Results within One Dataset*

To verify the benefits of the proposed method, it is compared with the unsupervised PCA-Net (U-PCA-Net) [12], the supervised PCA-Net (S-PCA-Net) [14] which achieve the state-of-arts performance on SAR image change detection. We also compare MASP-Net with the deep neural network (DNN) method [11] and CNN [13]. The patchsizes of U-PCA-Net and S-PCA-Net are set as 15×15. The ones of DNN and CNN are set as 11×11. These choices have been proven to be optimal through experiments by their developers. Among these methods, DNN and U-PCA-Net are unsupervised methods, while S-PCA-Net, CNN and MSSP-Net are supervised ones. The comparisons are performed on an individual dataset which means for the supervised methods, the training samples were drawn to train the models and the models were verified on the rest samples in the same datasets. All the compared methods take both the bi-tmeporal SAR images and their corresponding DI as the input data.

The visual comparison results are demonstrated in Figure 3. For the YR-A dataset, most methods can get clear changed regions, except that CNN gets some noisy spots. For the YR-B dataset, most methods miss the linear changed region, while both S-PCA-Net and MSSP-Net get more completed changed regions, especially the line at the bottom of the image. For Sendai-A and Sendai-B datasets, the backgrounds are so complex that the U-PCA-Net almost fails to work out the changed regions. CNN also mis-classifies several non-changed pixels as the changed ones. Instead, both S-PCA-Net and MSSP-Net can get more accurate changed regions.

Besides visual comparisons, we also compare the accuracies of change detection in Table 2. We can observe from the table that all compared methods obtain comparable accuracies on both YR-A and YR-B datasets. Instead, on the Sendai-A dataset, our proposed method MSSP-Net outperforms other compared methods, 1.89% and 4.31% better than CNN and S-PCA-Net, respectively. On Sendai-B dataset, both S-PCA-Net and CNN methods obtain comparable accuracies around 95.5%. Our proposed method MSSP-Net performs 0.54% better than S-PCA-Net and 0.61% than CNN.

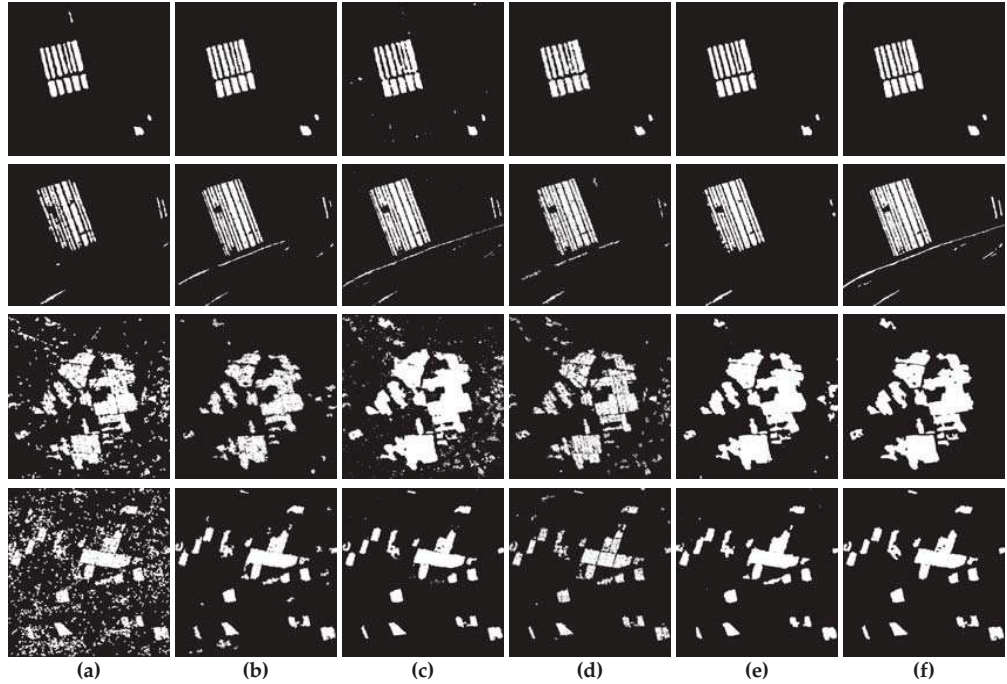

**Figure 3.** The visual experiment results of the compared methods: (**a**) unsupervised PCA-Net (U-PCA-Net), (**b**) supervised PCA-Net (S-PCA-Net), (**c**) CNN, (**d**) deep neural network (DNN), (**e**) MSSP-Net and (**f**) Reference map. From the above to bottom, they are the comparison results on YR-A, YR-B, Sendai-A and Sendai-B dataset, respectively.

Table 2. The change detection accuracies of compared methods.

| Datasets | U-PCA-Net | S-PCA-Net | CNN | DNN | MSSP-Net |
|---|---|---|---|---|---|
| YR-A | 98.93% | **99.98%** | 98.98% | 97.31% | 99.41% |
| YR-B | 95.70% | **97.92%** | 95.38% | 94.74% | 97.21% |
| Sendai-A | 90.30% | 90.75% | 93.17% | 89.64% | **95.06%** |
| Sendai-B | 86.63% | 96.00% | 95.93% | 90.30% | **96.54%** |

Furthermore, we also show more quantitative evaluation results in Figure 4. It is shown in Figure 4a that the MSSP-Net obtains a lower pFA on YR-B dataset, while a comparable pFA with CNN and DNN on Sendai-B dataset. Next, it is shown in Figure 4b that it can get the lowest pMA on the Sendai-A and Sendai-B datasets. It is shown that the pMA of U-PCA-Net is more larger than other compared methods and we set the upper bound as 0.1 to clearly compare the pMA among other methods. Overall, it is shown in Figure 4c that MSSP-Net obtains comparable kappa coefficients with CNN on the YR-B dataset and S-PCA-Net on the YR-A, respectively. On both Sendai-A and Sendai-B datasets with complex scenes, MSSP-Net gets the highest kappa among the comparison.

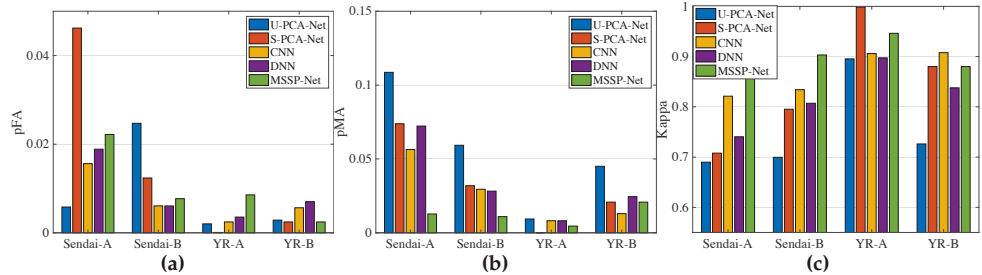

**Figure 4.** The comparisons of quantitative evaluations in terms of (**a**) false alarm (FA), (**b**) missed alarm (MA) and (**c**) Kappa coefficients.

*3.3. Experiment Results of Cross-dataset Change Detection*

It has been shown from the experiment results of the individual dataset change detection that MSSP-Net performs competitively with S-PCA-Net and CNN. In this section, to verify the generalization of the proposed method, we further compare the proposed method with two supervised change detection methods: S-PCA-Net and CNN on the cross-dataset change detection, where the networks trained on several datasets are applied to an unknown dataset. To achieve this, we conduct the comparisons through the leave-one-out validation, i.e., a dataset is alternatively selected as the testing dataset and the other three datasets are employed as the training datasets. Other configures are same as the above experiment.

The visual comparisons are shown in Figure 5. For YR-A dataset, both CNN and S-PCA-Net have obvious mis-classifications, while MSSP-Net gets more clear changed regions. For YR-B dataset, all three methods obtain missed detection with different levels. S-PCA-Net can obtain more completed changed regions, especially the line at the bottom of the image. For both Sendai-A and Sendai-B datasets, it is shown that S-PCA-Net produces noisy results. Instead, MSSP-Net obtains more accurate changed regions compared with other two methods.

Besides visual comparisons, we also compare the accuracies of change detection in Table 3. It is shown in the table that S-PCA-Net and our proposed method MSSP-Net obtain comparable accuracies on YR-A and YR-B datasets and S-PCA-Net gets a little higher accuracy than MSSP-Net. Instead, on Sendai-A dataset, MSSP-Net performs 2.12% better than CNN and 11.26% better than S-PCA-Net. On Sendai-B dataset, MSSP-Net performs 0.32% better than CNN and 6.79% better than S-PCA-Net.

Furthermore, more quantitative evaluations are shown in Figure 6. Similarly, the pMA and pFA of CNN are larger than other two methods, we set the upper bounds of pFA and pMA as 0.2 and 0.1 in panel a and b, respectively to clearly illustrate the comparison between other two methods. It is shown in Figure 6a that MSSP-Net obtains the lowest pFA on Sendai-A, Sendai-B and YR-B datasets, while in Figure 6b, S-PCA-Net gets lower pMA on YR-A, Sendai-A and Sendai-B datasets. Overall, it is shown in Figure 6c that MSSP-Net obtains comparable kappa values with other S-PCA-Net on YR-A and YR-B datasets. But it performs better other two methods on the challenging datasets: Sendai-A and Sendai-B with complex scenes with an obvious advantage in term of kappa.

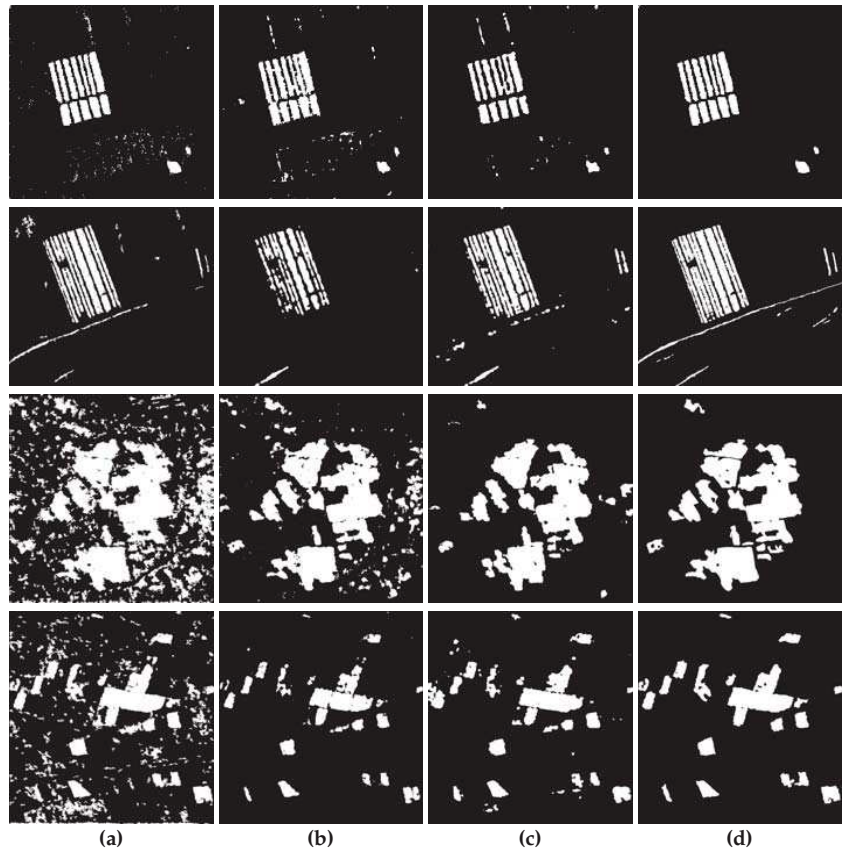

|     |     |     |     |
| :-: | :-: | :-: | :-: |
| (**a**) | (**b**) | (**c**) | (**d**) |

**Figure 5.** The visual comparison experiments: (**a**) S-PCA-Net, (**b**) CNN, (**c**) MSSP-Net and (**d**) Reference map. From the above to bottom, they are the comparison results on YR-A, YR-B, Sendai-A and Sendai-B dataset, respectively.

**Table 3.** The cross-dataset change detection accuracies of compared methods.

| Datasets | S-PCA-Net | CNN | MSSP-Net |
| --- | --- | --- | --- |
| YR-A | **99.45%** | 98.43% | 98.41% |
| YR-B | **97.64%** | 95.38% | 96.85% |
| Sendai-A | 82.72% | 91.86% | **93.98%** |
| Sendai-B | 89.87% | 95.38% | **95.66%** |

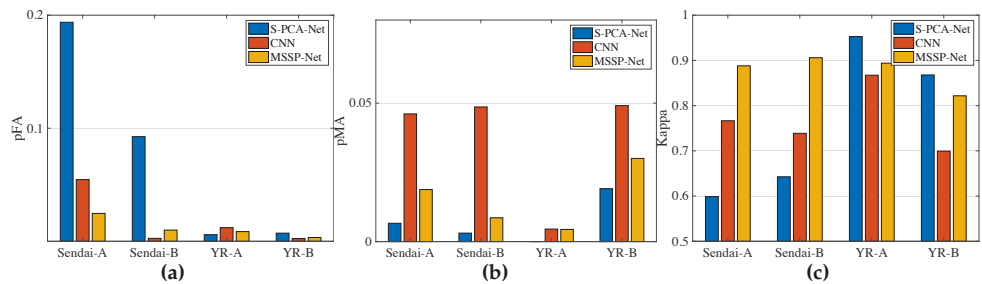

**Figure 6.** The comparisons of quantitative evaluations in terms of (**a**) FA, (**b**) MA and (**c**) Kappa coefficients.

*3.4. Computation Evaluation*

Overall, it is shown from above comparisons that MSSP-Net outperforms than other methods on Sendai-A and Sendai-B datasets. Furthermore, MSSP-Net shows great advantages on cross-dataset change detection, especially on two challenging datasets: Sendai-A and Sendai-B. Finally, we compare the executing times of S-PCA-Net, CNN and MSSP-Net in Table 4.

**Table 4.** The executing times of compared methods.

| Methods | S-PCA-Net | CNN | MSSP-Net |
| --- | --- | --- | --- |
| Training | 180 mins. | 30 mins. | 20 mins. |
| Testing | 5 mins. | 3 mins. | 20 secs. |

We evaluate the executing time on the crossing dataset change detection of Sendai-A and Sendai-B dataset. It is shown that compared with S-PCA-Net, a two-layer convolutional neural network, MSSP-Net is much more efficient and takes less time to train a model. It can exploit multiscale receptive field with a shallow network and be trained without the time-consuming singular value decomposition. Therefore, MSSP-Net is more efficient than S-PCA-Net. For the testing phase, due to the light network, our proposed MSSP-Net only takes 20 s to extract the changed region, much more efficiently than S-PCA-Net and CNN.

*3.5. Discussion*

So far, we verified our proposed method MSSP-Net on four sets of bitemporal SAR images and two scenarios of one-dataset and cross-dataset change detection. We compared our proposed method with state-of-art methods. The experiment results on four sets of bitemporal SAR images show that the proposed method can get more clear and completed changed regions than other compared methods, especially on the challenging datasets with complex scene. Furthermore, the proposed method has comparable results with our previous method S-PCA-Net on YR-A and YR-B dataset and outperforms other methods on Sendai-A and Sendai-B dataset in terms of kappa coefficients and MA. The comparisons on cross-dataset show that our method get lower pFA than other methods on all four datasets. In term of kappa coefficient, our method gets comparable results on YR-A and YR-B dataset and performs best on both Sendai-A and Sendai-B dataset.

Furthermore, in term of computation efficiency, our method is more efficient on both training and testing phase. Especially, MSSP-Net is much more efficient than S-PCA-Net, which achieved best performance on YR-A and YR-B dataset. Especially compared with CNN, the sequential version of MSSP-Net, our proposed method performs better in terms of kappa coefficient and computation efficiency (both training and testing phase).

**4. Conclusions**

In this paper, we developed a simple and efficient deep learning method for bitemporal SAR image change detection. In this method, we designed a multiscale spatial pooling layer that exploits the spatial context at various scales parallel without the depth increase of the network, which is the most significant difference from the conventional convolutional neural network. We applied our proposed method to the cross-dataset bitemporal SAR image change detection to verify its generalization and compare with several deep learning methods recently proposed for the SAR image change detection. In one-dataset change detection experiment, the proposed method show the great advantages on both Sendai-A and Sendai-B datasets with complex scenes. Furthermore, in the cross-dataset experiment, the proposed MSSP-Net obtained comparable results with S-PCA-Net on YR-B dataset and better performance than other methods on Sendai-A and Sendai-B dataset. Also, the proposed method is more efficient than other compared supervised methods. In the future, we will focus on how to optimize the network and make it more efficient.

**Author Contributions:** J.-W.C. designed the project and wrote the manuscript; R.W. and B.L. designed the experiments and analyzed the data; F.D. and J.Z. performed the experiments. L.J. revised the paper. All authors have read and agreed to the published version of the manuscript

**Funding:** This work was supported by the State Key Program of National Natural Science of China (No. 61836009), the National Natural Science Foundation of China (No. 61701361, 61806154) and the Open Fund of Key Laboratory of Intelligent Perception and Image Understanding of Ministry of Education, Xidian University (Grant No. IPIU2019006).

**Conflicts of Interest:** The authors declare no conflict of interest.